\documentclass[letterpaper, 10 pt, conference]{ieeeconf}  

\IEEEoverridecommandlockouts                              
\usepackage{kotex}
\usepackage{graphicx}
\usepackage{array}
\usepackage{tabularx} 
\usepackage{booktabs}  
\usepackage{multirow}   
\usepackage[table]{xcolor}
\usepackage{makecell}
\usepackage{soul}
\usepackage[framemethod=tikz]{mdframed}
\usepackage{amsmath, amssymb}
\usepackage[
  style=ieee,
  backend=biber,
  sorting=none,   
  sortcites=true,
  maxnames=3, minnames=1,    
  giveninits=true,   
  doi=false, url=false, isbn=false, eprint=false
]{biblatex}
\usepackage{etoolbox}
\AtBeginBibliography{\footnotesize}

\usepackage[capitalize,noabbrev]{cleveref}

\title{\LARGE \bf
ThermoAct: Thermal-Aware Vision-Language-Action Models \\for Robotic Perception and Decision-Making

}
\author{Young-Chae Son$^{\dagger}$, Dae-Kwan Ko$^{\dagger}$, Yoon-Ji Choi, Soo-Chul Lim*%
\thanks{Manuscript received: October 1, 2025; Revised: December 23, 2025; Accepted: February 3, 2026.}%
\thanks{This paper was recommended for publication by Editor Markus Vincze upon evaluation of the Associate Editor and Reviewers comments. This work was supported by the National Research Foundation of Korea (NRF) grant funded by the Korea government (MSIT) (RS-2025-00562981, RS-2025-25433409). (Young-Chae Son and Dae-Kwan Ko contributed equally to this work.) (Corresponding author: Soo-Chul Lim.)}%
\thanks{The authors are with the Department of Mechanical, Robotics and Energy Engineering, Dongguk University, Seoul 04620, South Korea. email: sonyc@dgu.ac.kr; daekwanko@dgu.ac.kr; choiyj@dgu.ac.kr; limsc@dongguk.edu}%
}

\begin{document}

\maketitle
\thispagestyle{empty}
\pagestyle{empty}

\thispagestyle{empty}

\begin{abstract}
In recent human-robot collaboration environments, there is a growing focus on integrating diverse sensor data beyond visual information to enable safer and more intelligent task execution. Although thermal data can be crucial for enhancing robot safety and operational efficiency, its integration has been relatively overlooked in prior research. This paper proposes a novel Vision-Language-Action (VLA) framework that incorporates thermal information for robot task execution. The proposed system leverages a Vision-Language Model (VLM) as a high-level planner to interpret complex natural language commands and decompose them into simpler sub-tasks. This approach facilitates efficient data collection and robust reasoning for complex operations. Unlike conventional methods that rely solely on visual data, our approach integrates thermal information, enabling the robot to perceive physical properties and proactively ensure environmental safety. Experimental results from real-world task scenarios validate the feasibility of our proposed framework, suggesting its potential to enhance task success rates and safety compared to existing vision-based systems.
Project page: \url{https://youngchae912.github.io/ThermoAct/}
\end{abstract}

\section{Introduction}
Recently, Vision-Language-Action (VLA) models have rapidly emerged as a prominent framework for realizing this integrated capability, demonstrating their potential to enable robots to comprehend language commands grounded in visual information and execute a diverse array of real-world tasks\cite{brohan2023rt,team2024octo,kim2024openvla,sapkota2025vision,ma2024survey}. Early VLA studies predominantly utilized images acquired from RGB cameras as the primary visual input\cite{brohan2022rt,kim2024openvla}. However, in real-world environments, relying solely on 2D image information is often insufficient for executing a wide range of tasks. More intelligent and safer interactions become possible only when a robot can integrate and process diverse sensory information, such as the position and distance of objects, tactile feedback, and sound. Consequently, recent VLA research has been expanding to incorporate multimodal information from various sensory modalities, including depth, tactile, and audio\cite{jones2025beyond,huang2025tactile,zhao2025vlas}. Contemporary VLA models have demonstrated the ability to perform a variety of complex tasks, such as folding clothes and clearing cluttered tables. Despite these advancements, the physical property of temperature remains largely overlooked. Consequently, current systems are unable to execute a seemingly simple command like ``pick up the coldest coke,'' as they can visually recognize the object but cannot perceive its thermal state to make the correct selection. For instance, in an environment with a hot hair straightener or an active induction cooktop, a robot with thermal perception can proactively detect and mitigate potential fire hazards. In contrast, a robot lacking this sense would merely identify the object's presence, failing to comprehend the situational risk.

We propose \textbf{ThermoAct}, a VLA-based approach aimed at handling these challenges. ThermoAct is designed to integrate thermal information obtained from a thermal camera into the task planning system, enabling the robot to perform temperature-based decision-making. This work demonstrates that by integrating thermal imagery with a VLA, a robot can make more intelligent and safer choices. In particular, tasks such as selecting the `coldest' object or identifying `hot and hazardous' situations require a level of abstract reasoning that extends beyond simple image-to-action mapping. Training an end-to-end VLA model directly for such complex reasoning is inefficient, especially given the current scarcity of large-scale thermal datasets. Therefore, this study shows that a hierarchical architecture, which uses a Vision-Language Model (VLM) as a high-level planner, can be a viable approach to overcome this problem. This structure effectively decomposes a complex main task into simpler sub-tasks that the VLA can readily learn, enabling stable execution even in long-horizon sequences. In this paper, we apply this hierarchical framework to real-world scenarios, experimentally analyzing and verifying the impact of thermal information on the robot's execution performance, stability, and safety.

\section{Related Works}
\subsection{Integrative Approaches: VLMs and VLAs}
Vision-Language Models (VLMs), trained on large-scale multimodal data, possess extensive prior knowledge and remarkable reasoning capabilities. Previous studies have shown that such knowledge enables robots to recognize objects and perform tasks without additional training ~\cite{kwon2024language,singh2024malmm}. Moreover, VLM reasoning has been leveraged for robot task planning, improving intelligent action generation and situational adaptability \cite{guo2024doremi,kannan2024smart}.
An early example of applying large-scale data-driven learning to robotics is RT-1, the first large-scale robotics transformer model, which demonstrated the ability to perform diverse tasks with a high success rate \cite{brohan2022rt}. Subsequent research has focused on developing capabilities that are not confined to specific environments and can handle novel tasks and situations, leading to the advancement of Vision-Language-Action (VLA) models \cite{team2024octo,kim2024openvla}. VLAs excel at generating robot actions by comprehensively understanding natural language instructions and visual information. Recently proposed models like $\pi_0$ \cite{black2024pi_0} and $\pi_0.5$ \cite{pi05} have shown that they can perform various tasks with only simple fine-tuning. While VLAs have the advantage of learning natural behaviors from small amounts of data \cite{kim2024openvla,black2024pi_0}, they exhibit performance degradation in complex or long-horizon tasks \cite{hu2023look}. To overcome this limitation, research is being conducted to integrate the high-level reasoning of Vision-Language Models (VLMs) with the low-level actions of VLAs by adding a VLM to the top of the framework, thereby achieving deep reasoning and natural actions \cite{yang2025agentic,shi2025hi,hu2023look}. ViLa \cite{hu2023look} overcomes these limitations by directly integrating the visual understanding capabilities of GPT-4V into the planning process, showing that this approach can particularly reduce comprehension errors. Inspired by the structure of the human brain, Agentic Robot \cite{yang2025agentic} proposes a framework that unifies the active reasoning and actions of VLM-based robots.

\subsection{Combining Multimodal dataset and VLAs}
It is well-established that using visual information or integrating additional sensor data aids robot manipulation \cite{ko2023vision,lee2024dextouch}. However, existing VLA models have predominantly focused on the visual modality and have been applied to relatively simple tasks. To address this, recent studies have proposed integrated frameworks that incorporate additional multimodal information, enabling the execution of tasks that are difficult for conventional VLA models. For example, \textcite{zhen20243d} developed an LLM with strong 3D perception capabilities that directly processes 3D point clouds instead of 2D images, demonstrating an understanding of 3D space. ECoT \cite{zawalski2024robotic} utilizes depth information to improve CoT-based reasoning for real-world manipulation tasks by leveraging an understanding of 3D spatial structures. ForceVLA \cite{yu2025forcevla} integrates 6D force data from the robot's end-effector, showing that directly measuring interaction forces can improve success rates in contact-rich tasks. TLA \cite{hao2025tla} was trained using a tactile sensor and demonstrated strong performance in contact-rich tasks. Furthermore, VTLA \cite{zhang2025vtla} combined images from a wrist-mounted camera with the GelStereo sensor, training on a dataset of Vision-Tactile-Action-Prompt pairs, and achieved higher success rates than the original TLA. Despite these advancements in incorporating force and sound, research on understanding physical interactions related to thermal properties remains scarce. This study investigates methods for performing tasks involving temperature by integrating thermal information into a Vision-Language-based model.

\subsection{Thermal Sensing in Robotic Perception and Planning}
Thermal information is valuable for enabling robotic systems to perceive their surroundings with greater precision and to assess the state of humans or objects. While conventional RGB-based sensing can be limited by factors such as light and smoke, \textcite{cruz2021autonomous} demonstrated that by mounting a thermal camera on a quadruped robot, human detection is possible even in low-illuminance search and rescue environments. More recently, there have been attempts to leverage Multimodal Large Language Models to enhance object detection and scene understanding from thermal images \cite{ashqar2024leveraging}. \textcite{barros2022cost} proposed a system where a collaborative robot uses a thermal camera to detect human presence, allowing it to automatically switch to a safe mode and reduce the risk of collision with workers. Research has also been conducted on classifying the state or type of an object by attaching temperature sensors to a robot gripper \cite{park2023object, osawa2022material, im2025simultaneous}. Notably, \textcite{im2025simultaneous} proposed a method that first estimates an object's approximate temperature using an IR camera and then measures the precise surface temperature through contact with a sensor-equipped gripper.

Previous studies have demonstrated the applicability of thermal information across a range of tasks. Building upon this, our work integrates thermal information into a hierarchical Vision-Language-Action framework, enabling temperature-aware general task execution and decision-making in real-world environments.

\begin{figure*}[!t]
    \centering
    \includegraphics[width=\textwidth]{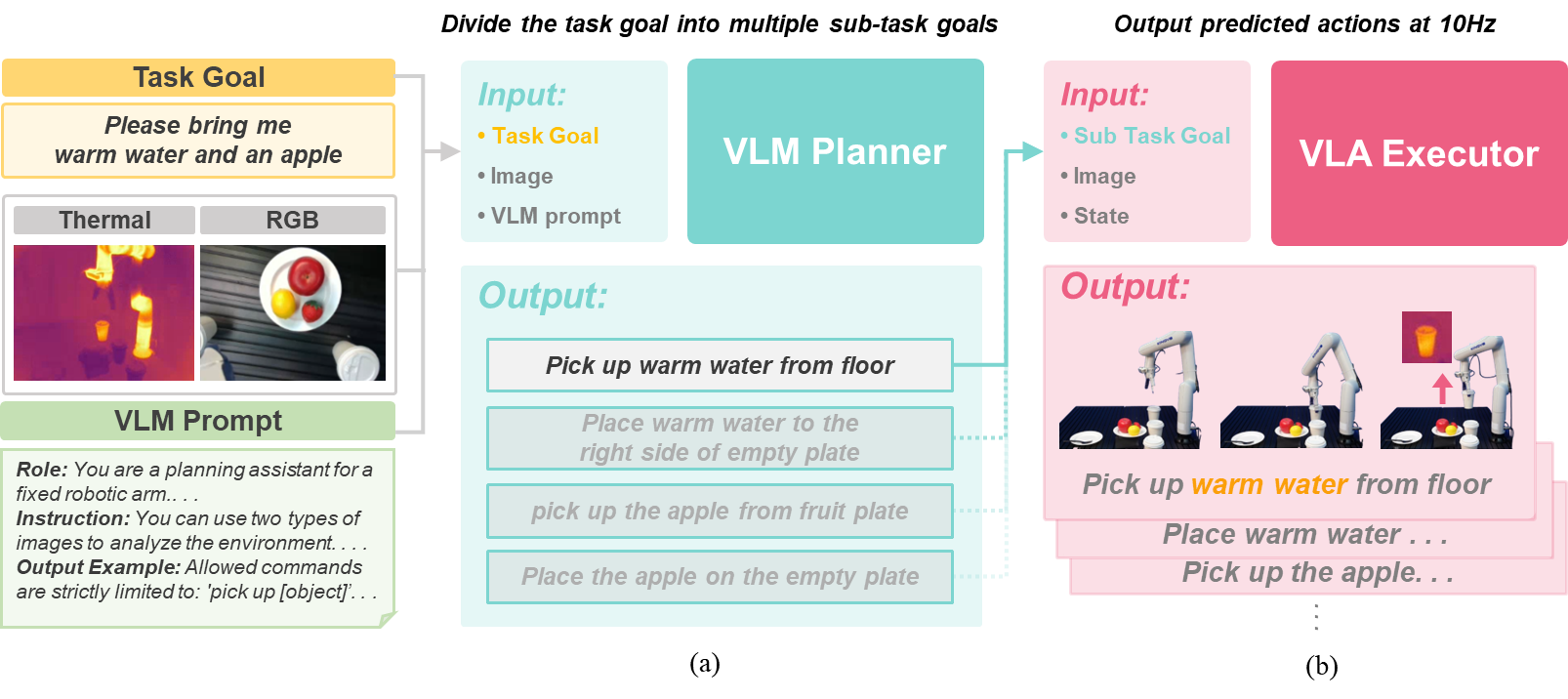}
    \caption{We propose \textbf{ThermoAct}. (a) illustrates a VLM Planner that decomposes a \textit{high-level user instruction} into specific \textit{sub-task descriptions}. (b) depicts a VLA Executor that receives these descriptions as input prompts to predict low-level actions. By leveraging temperature cues from thermal imaging, ThermoAct is able to perform temperature-aware tasks beyond existing approaches.}
    \label{fig:method-framework}
\end{figure*}

\section{Method}

\section{ThermoAct Architecture}
The ThermoAct framework proposed in this study consists of a Vision-Language Model (VLM) that performs reasoning and planning based on user commands and environmental information, and a Vision-Language-Action (VLA) module that executes robot control commands based on this plan. The VLM takes visual inputs, including thermal data, and a natural language instruction to generate a low-level action plan tailored to the situation. Subsequently, the VLA module controls the robot in real-time based on the decomposed plan and the corresponding inputs.

\subsubsection*{\textbf{Experiment Setups}}
The experiments were conducted using a 7-DoF Kinova Gen3 Lite robot. The camera setup included two RGB-D cameras (Realsense D435, Realsense L515) and one thermal camera (ThermoEye TMC256B). For the RGB-only baseline tasks, we used the D435 as the wrist camera and the L515's RGB stream as the external camera for data collection, training, and evaluation. For our proposed model, TaVLA, the D435 was used as the wrist camera, while the TMC256B was used as the external camera for training and evaluation. All experiments were performed within a temperature range of 20-35\,$^\circ$C The average recorded indoor temperature during the experiments was 21.5\,$^\circ$C.

\subsubsection*{\textbf{Dataset for VLA Executor}}
To train the VLA Executor, we collected 50 demonstrations per task and applied LoRA\cite{hu2022lora}-based fine-tuning to adapt the model efficiently with limited data. To enhance data diversity, we varied the position and orientation of the objects during the data collection process. The dataset is composed of four components: \emph{State}, \emph{Action}, \emph{Image}, and \emph{Task prompt}.

The \emph{State} is a 7-dimensional vector comprising six joint angles and the gripper value. The \emph{Action} is an 8-dimensional vector that includes six target joint angles, a gripper control value, and a termination flag. The done flag was defined during data collection. The \emph{Image} component was collected via the wrist and external cameras. All cameras were synchronized at a frame rate of 15 Hz. The RGB images were captured using Realsense D435 and L515 cameras at a resolution of 640$\times$480. The detailed processing pipeline for the thermal image component is described in the following subsection.
The \emph{task prompt} consists of a natural language sub-task description generated by the VLM Planner

\subsubsection*{\textbf{Dataset containing thermal information}}
The dataset in this study integrates visual and thermal information to enable the robot to perform thermal-related tasks. All data were collected and synchronized at a rate of 15 Hz. Taking Task 1 as an example, the dataset consists of 50 episodes, totaling 49,343 frames ($\approx$ 55 minutes).  Within this task, sub-tasks that directly utilize thermal information, such as ``pick up warm water from the floor,'' account for 14,767 frames ($\approx$ 16 minutes), representing about 30\% of the total operation time. To ensure that the Vision-Language-Action (VLA) model can effectively learn from the raw thermal data ($256 \times 192$), the data is converted into RGB images through the following preprocessing steps. First, the raw data is linearly normalized within a target indoor temperature range of 20$^\circ$C to 35$^\circ$C. 
Finally, the transformed values are quantized into 8-bit grayscale intensities and mapped to the INFERNO pseudocolor palette. This mapping assigns dark purple to lower temperatures and bright yellowish-white to higher temperatures, providing distinct visual features that the VLA model can effectively encode.

\subsubsection*{\textbf{VLA Executor}}
The inputs and outputs of the VLA Executor are illustrated in Fig.~\ref{fig:method-framework} (b). Our VLA model is based on the $\pi_0$ \cite{black2024pi_0} and predicts actions by taking the low-level task from the VLM Planner and real-time camera images as input. At each control cycle (10 Hz), the model receives a thermal image (256$\times256\times$3), a wrist RGB image (256$\times256\times$3), a 7-dimensional robot state vector, and the natural language task description prompt generated by the VLM Planner. At a given timestep $t$, the robot's observation is composed of the fixed external camera input $V_t^e$, the wrist camera input $V_t^w$, the robot state vector $s_t \in \mathbb{R}^7$ (including the gripper state), and the task description $\tau$:
\begin{equation}
o_t = \{ V_t^e, V_t^w, s_t, \tau \}.
\end{equation}

As output, the model generates an action at each timestep, which includes six target joint angles, a gripper control value, and a done flag:
\begin{equation}
a_t = \big[ \theta_{t}^{(1)}, \theta_{t}^{(2)}, \dots, \theta_{t}^{(6)}, g_t, d_t \big].
\end{equation}

The VLA autonomously predicts $d_t$ to signal sub-task completion. This triggers a transition to the next sub-task, where the VLA receives the subsequent VLM instruction as a new prompt, continuing until the full plan is executed.

\begin{figure*}[t]
   \includegraphics[width=\textwidth]{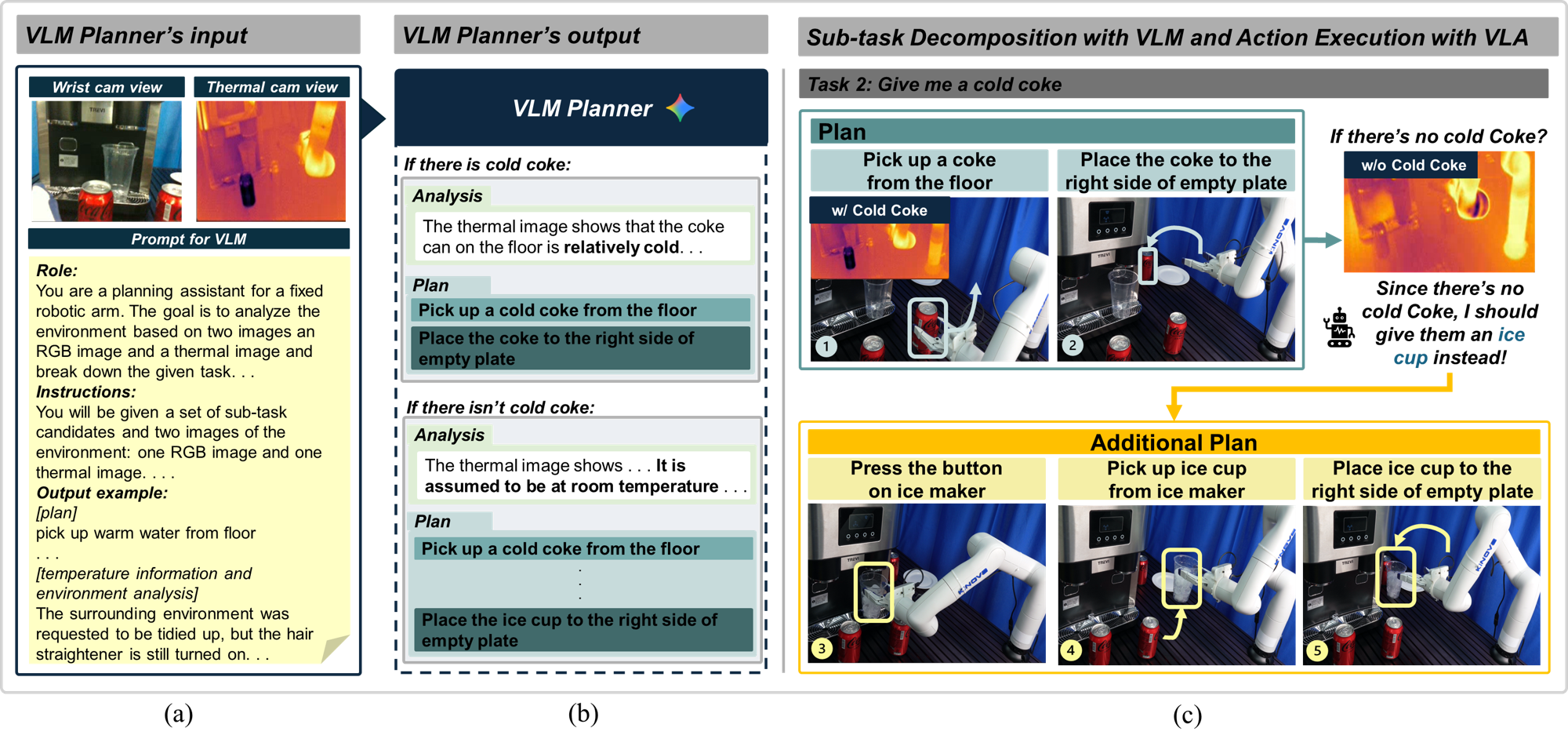}
    \caption{\textbf{Hierarchical Collaboration between VLM Planner and VLA Executor.} (a) The VLM Planner receives RGB-Thermal images and a structured guideline prompt containing role definitions and output examples. (b) Based on the thermal information, the VLM analyzes the environment context and decomposes the instruction into executable sub-tasks. (c) Sub-task Decomposition with VLM and Action Execution with VLA.}
   \label{fig:give_me_a_cold_cola}
\end{figure*}

\subsubsection*{\textbf{VLM Planner}}
The inputs and outputs of the VLM Planner are illustrated in Fig.~\ref{fig:method-framework} (a). We utilized Gemini 2.0 Flash \cite{GoogleBlogGemini2} as the planner, which operates at the highest level of the ThermoAct architecture. It receives a user-defined task and decomposes it into multiple sub-plans for execution. As input, the planner takes the desired user task, a thermal image (256$\times256\times$3), a wrist RGB image(256$\times256\times$3), and a guideline prompt. To ensure correct reasoning and response generation, this prompt is structured with a Role, Environment instructions, an Output format, and an Output example. As output, the planner provides an analysis of the current environment and a decomposed low-level task plan that can be executed by the VLA.

\section{Experiment}
In this paper, we conducted a total of three experiments to evaluate the potential for intelligent robot behavior using thermal information by combining a thermal camera, a Vision-Language Model (VLM), and a Vision-Language-Action (VLA) model. To prevent biased results, the position and orientation of the objects were randomized in each test.

\subsection{Task Descriptions}
All tasks were designed to require the successful completion of not only temperature-aware sub-tasks but also everyday sub-tasks (e.g., clearing unused cables, placing an apple on a plate). 
The sub-tasks decomposed by the VLM Planner follow a standardized format, aligning with protocols used in recent VLM-based robotic planning research~\mbox{\cite{yang2025agentic,shi2025hi}}.

Task 1 to Task 3 were designed to evaluate whether the robot could act more intelligently by utilizing thermal information in daily scenarios, such as handing over a cup of warm water or giving a cold can of soda. 
Task 4 and Task 5 was designed to verify the utility of thermal information in safety-related situations, such as picking up an overheated battery, turning off a hot hair straightener, and organizing the nearby power strip.

\textbf{Task 1. Please bring me warm water and an apple:}
The robot must identify and hand over the warm cup from a selection of several cups, and then select an apple from a variety of fruits to deliver. To introduce variation, half of the trials involved placing all fruits onto a single plate, while the other half involved placing them on separate plates. The temperature of the warm cup was set to a range of 28-32\,$^\circ$C. 
\begin{figure*}[t]
   \includegraphics[width=\textwidth]{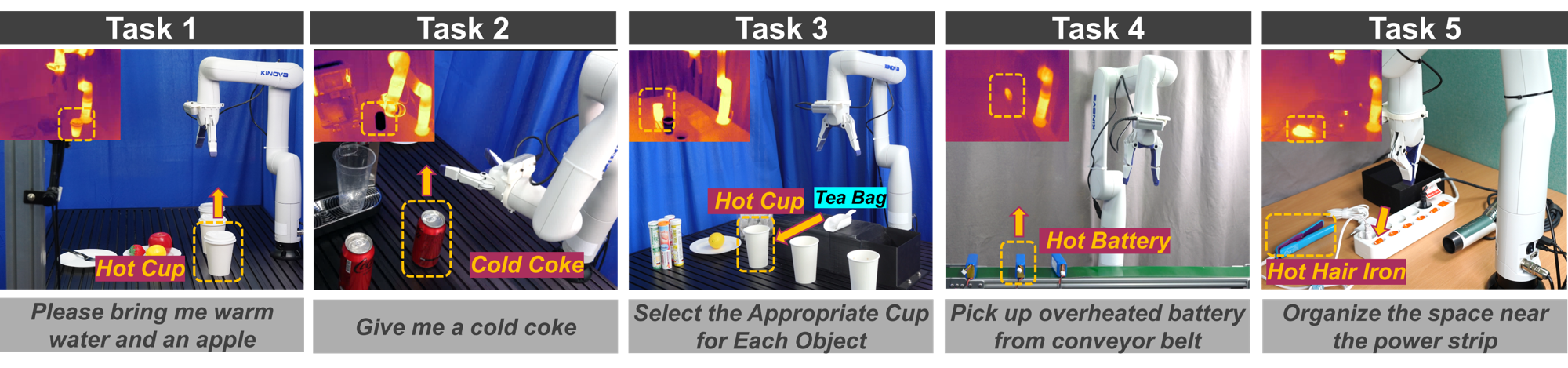}
    \caption{The figure shows five main task environments (Tasks 1--5), with the actual thermal input images displayed above each task. Tasks 1--3 correspond to daily-life manipulation tasks, while Tasks 4--5 focus on safety-related scenarios.}
    \label{fig:task_description}
\end{figure*}

\begin{figure}[t]
   \includegraphics[width=\columnwidth]{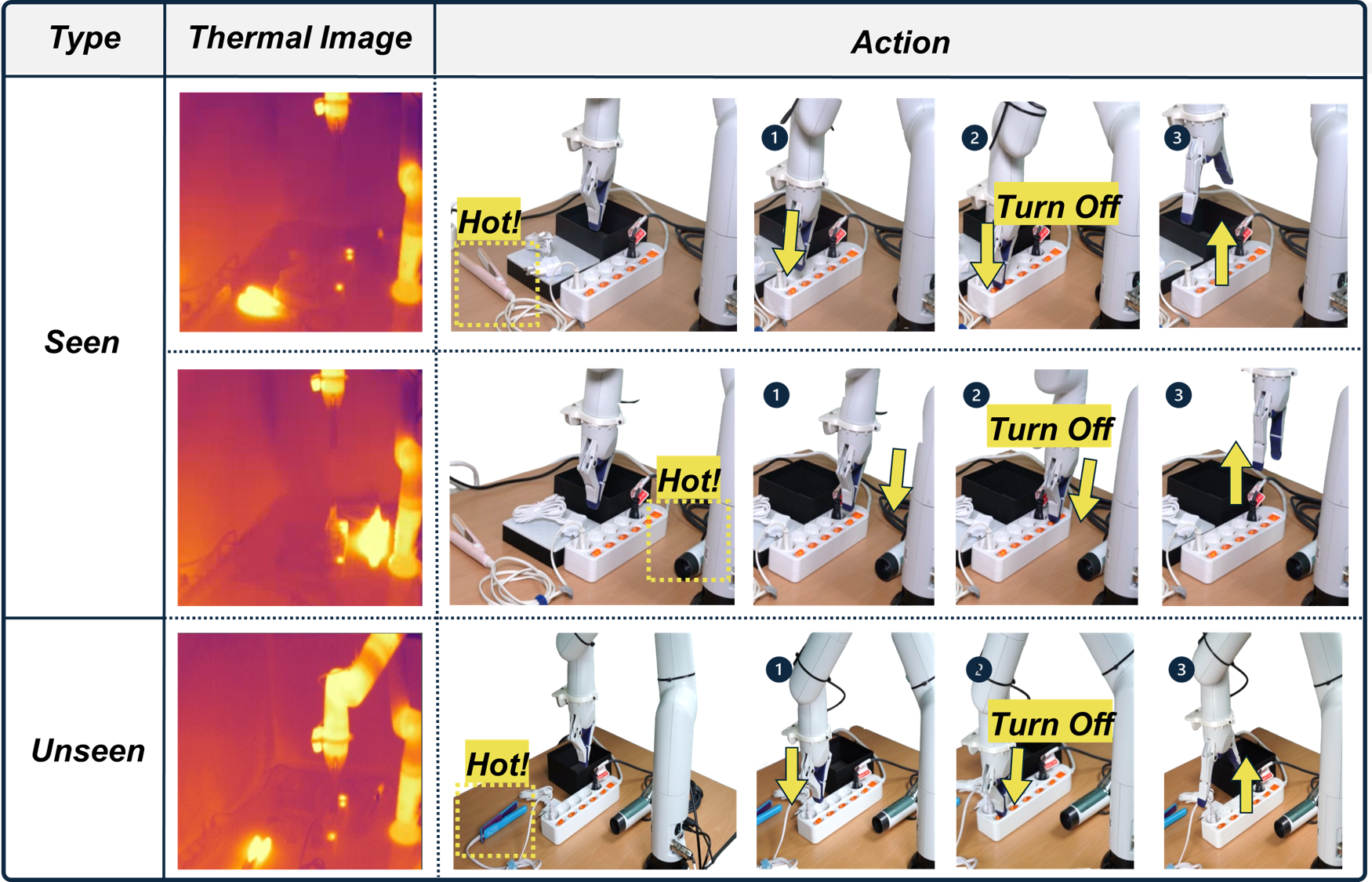}
   \caption{Process of Task 5, where the system turns off the heated hair straightener and successfully generalizes to unseen ones using the learned data.}
   \label{fig:multimodal_example}
\end{figure}
\textbf{Task 2. Give me a cold Coke:}The robot determines whether a cold can of Coke is present among the available options. As shown in Fig.~\ref{fig:give_me_a_cold_cola}, the VLM Planner generates a corresponding plan based on this assessment. If a cold Coke is available, the robot delivers it directly. If not, it uses an ice maker to fill an empty cup with ice and serves it with a can of Coke. Consequently, the experiment was conducted for a total of 10 trials, with 5 trials for each condition. The temperature of the cold Coke was set to a range of 15-18\,$^\circ$C. 

\textbf{Task 3. Select the Appropriate Cup for Each Object:}
The robot is presented with three cups containing warm water, cold water, and coke, respectively, along with a single, randomly placed target object. The robot must utilize visual recognition and VLM-based reasoning to identify the object's properties and select the appropriate cup. Specifically, the robot is tasked with placing tea bags, which require warm water, into warm cups and placing lemons into cups containing Coke. A total of 10 trials were conducted. 

\textbf{Task 4. pick up overheatted battery from conveyor belt:}
The robot is tasked with detecting and picking up an overheated battery from a set of lithium batteries moving along a conveyor belt. Three batteries travel at a constant speed, and when one of them exhibits an anomalous temperature profile compared to the others, the robot lifts the overheated battery.

\textbf{Task 5. Organize the space near the power strip:} This task evaluates the robot's ability to detect and respond to hazards using thermal information. The robot must clear unnecessary cables from around a power strip and, if it detects an active appliance like a hot hair straightener, turn it off. Fig.~\ref{fig:multimodal_example} illustrates the process of turning off the hair straightener as an example.

\subsection{Experiment in Real-World}

\subsubsection{Experiment 1: Thermal Perception and Action Accuracy}
This experiment evaluated how accurately the robot can perform temperature-based actions using a thermal camera. We also verified whether it is possible to learn correct behaviors by fine-tuning with a small amount of thermal data, and we quantified the performance difference compared to a conventional RGB-RGB model. We fine-tuned the models on four distinct tasks using 30, 50, and 70 episodes, respectively. The success rate, based on task completion, was used as the primary metric.

\begin{itemize}
\item Pick apple from various fruits (RGB-RGB)
\item Pick apple from various fruits (RGB-T)
\item Pick the warmest cup (RGB-T)
\item Turn off the active hair straightener (RGB-T)
\end{itemize}

\subsubsection{Experiment 2: Input Modality Comparison}
In this experiment, we quantitatively analyzed the impact of different input modality configurations on performance. To this end, we configured the following two models:

\begin{itemize}
\item \textbf{RGB-RGB:} This model follows the conventional VLA approach, using RGB images from both the wrist and external cameras, along with the state, action, and prompt as inputs. It was fine-tuned on 50 demonstration episodes.
\item \textbf{Ours (RGB-T):} This is our proposed method, where the model is trained by combining the RGB image from the wrist camera with the thermal image from the external camera.
\end{itemize}

\begin{table}[t]
  \centering
  \caption{Performance of Fine-Tuned Tasks by Episode}
  \label{tab:rgb_vs_thermal_avg_sd}
  \resizebox{\columnwidth}{!}{%
  \begin{tabular}{lccc}
    \toprule
    \textbf{Method} & \textbf{30 episode} & \textbf{50 episode} & \textbf{70 episode} \\
    \midrule
     RGB-RGB & 60.0\% & 80.0\% & 90.0\% \\
    \textbf{Ours(RGB-T)} & \textbf{43.3 $\pm$ 11.5\%} & \textbf{83.3 $\pm$ 5.8\%} & \textbf{86.7 $\pm$ 5.8\%} \\
    \bottomrule
  \end{tabular}%
  }

\end{table}

\begin{table*}[t]
    \centering
        \begin{minipage}{0.95\textwidth}
            \centering
            \caption{\textbf{Comprehensive Success Rates Comparison.} This table integrates sub-task performances and overall averages. All success rates are reported based on $N=10$ independent trials per sub-task.}
            \label{tab:integrated_results}
            \renewcommand{\arraystretch}{1.2}
            \resizebox{\linewidth}{!}{%
                \begin{tabular}{llccc}
                \toprule
                \multirow{2}{*}{\textbf{Task Scenario}} & \multirow{2}{*}{\textbf{Sub-task}} & \multicolumn{3}{c}{\textbf{Success Rate (\%)}} \\ \cmidrule(l){3-5} 
                 &  & \textbf{FlatVLA} & \textbf{RGB-RGB} & \textbf{Ours (RGB-T)} \\ \midrule

                \multicolumn{5}{l}{\textit{\textbf{Task 1: Bring warm water and an apple}}} \\ 
                 & pick up [warm water] from [floor] & - & 40 & \textbf{90} \\
                 & place [warm water] to the [right side] of [empty plate] & - & 100 & 90 \\
                 & pick up [an apple] from [fruit plate] & - & 70 & 70 \\
                 & place [an apple] on the [empty plate] & - & 80 & 70 \\ 
                \rowcolor{gray!15} \textbf{Task 1 Average} &  & \textbf{0.0} & \textbf{72.5 $\pm$ 25.0} & \textbf{80.0 $\pm$ 11.5} \\ \midrule

                \multicolumn{5}{l}{\textit{\textbf{Task 2: Give me a cold Coke}}} \\ 
                 & pick up [coke] from [floor] & - & 70 & \textbf{90} \\
                 & place [coke] to the [right side] of [empty plate] & - & 90 & 80 \\ 
                 & press [the button] on [ice maker] & - & 60 & 80 \\
                 & pick up [ice cup] from [ice maker] & - & 60 & 60 \\ 
                 & place [ice cup] to the [right side] of [empty plate] & - & 60 & 60 \\ 
                \rowcolor{gray!15} \textbf{Task 2 Average} &  & \textbf{10.0} & \textbf{68.0 $\pm$ 13.0} & \textbf{74.0 $\pm$ 13.4} \\ \midrule

                \multicolumn{5}{l}{\textit{\textbf{Task 3: Select the Appropriate Cup for Each Object}}} \\ 
                 & pick up [the scoop] from [floor] & - & 100 & 100 \\
                 & pour [scoop] into the [coke/hot water] & - & 40 & \textbf{60} \\
                \rowcolor{gray!15} \textbf{Task 3 Average} &  & \textbf{0.0} & \textbf{70.0 $\pm$ 42.4} & \textbf{80.0 $\pm$ 28.3} \\ \midrule

                \multicolumn{5}{l}{\textit{\textbf{Task 4: Pick up overheated battery from conveyor belt}}} \\ 
                 & pick up [overheated battery] & 80 & 30 & \textbf{80} \\
                \rowcolor{gray!15} \textbf{Task 4 Average} &  & \textbf{80.0} & \textbf{30.0 $\pm$ 0.0} & \textbf{80.0 $\pm$ 0.0} \\ \midrule
               
                \multicolumn{5}{l}{\textit{\textbf{Task 5: Organize space near power strip}}} \\ 
                 & turn off [hair straightener] & - & 30 & \textbf{90} \\
                 & pick up [unplugged wire] from [floor] & - & 70 & 60 \\
                 & place [unplugged wire] to [power strip] & - & 70 & 60 \\ 
                \rowcolor{gray!15} \textbf{Task 5 Average} &  & \textbf{0.0} & \textbf{56.7 $\pm$ 23.1} & \textbf{70.0 $\pm$ 17.3} \\ \midrule

                 \textbf{Overall Average} & \textbf{(All Tasks 1-5)} & \textbf{18.0 $\pm$ 34.9} & \textbf{59.4 $\pm$ 17.1} & \textbf{76.8 $\pm$ 4.6} \\ \bottomrule
                \end{tabular}%
            }
            \vspace{5pt} 
        \end{minipage}%
\end{table*}
\subsubsection{Experiment 3: Planning via VLM vs. Flat VLA Models}
In this experiment, we compared two distinct approaches. The first is a hierarchical method, in which a Visual Language Model (VLM) is employed to semantically decompose a complex task into a series of sub-tasks, which are then executed sequentially by a Vision-Language-Action (VLA) model. The second is a single end-to-end learning approach, where a flat VLA model directly learns to perform the entire task without any intermediate decomposition. A major limitation in research involving thermal camera data is the lack of large-scale pre-training datasets. Consequently, a single VLA model must be trained on relatively limited data, which can hinder its generalization performance. In contrast, the hierarchical approach can potentially reduce data requirements by breaking down tasks into smaller units and may enable more stable performance in complex scenarios. The purpose of this experiment is not only to compare the performance of the two methods, but also to explore which structural approach is relatively more advantageous under data-constrained conditions. To this end, we quantitatively evaluated differences in task success rates between the hierarchical method and the flat VLA model.

\section{Result}
\subsubsection{Experiment 1: Thermal Perception and Action Accuracy}
The results for tasks fine-tuned on 30, 50, and 70 episodes are presented in Table~\ref{tab:rgb_vs_thermal_avg_sd}.

We evaluated the performance of our proposed model, ThermoAct (RGB-T), on a set of sub-tasks: a non-thermal task (``pick the apple'') and two thermal-aware tasks (``pick the warmest cup of water'' and ``turn off the active hair straightener''). The overall average success rate for ThermoAct increased from 43.3\% ± 11.5 to 83.3\% ± 5.8, and finally to 86.7\% ± 5.8, showing a stabilizing trend as training progressed. For the ``pick apple'' task, the initial success rate of the RGB-RGB model (60\%) was higher than that of the RGB-T model (50\%), but both models converged to 80\% at the 50-episode mark. The ``pick the warmest cup'' task started with a success rate of only 30\% at 30 episodes but showed a significant improvement to 90\% after 50 episodes. This suggests that with additional data, the model learned to effectively detect and prioritize thermal cues. The task of turning off the active hair straightener began with a 50\% success rate at 30 episodes and gradually improved, ultimately reaching 90\%. These results indicate that using thermal camera images as input does not cause a significant performance degradation compared to fine-tuning with only RGB data. Furthermore, both models demonstrated improved overall accuracy and stability as the amount of fine-tuning data increased. The experimental results showed a tendency for performance to converge or stabilize after improvement by the 50-episode mark in most tasks. Therefore, considering training efficiency, such as data collection costs and computational resources, we present the 50-episode results as the primary findings in this study.

\subsubsection{Experiment 2: Input Modality Comparison}
In this experiment, we evaluated the performance of the proposed RGB-T model across five main tasks. As presented in Table II, success rates are reported based on 10 independent trials for each sub-task. To ensure a fair comparison, the execution sequence of sub-tasks was maintained consistently across all input modalities. Our objective was to investigate whether the proposed model maintains robust performance even in sub-tasks irrelevant to thermal properties. Therefore, the average signifies the mean success rate across individual sub-tasks, distinct from the end-to-end task completion rate. The results demonstrate that the proposed model achieved a high average success rate, showing distinct performance improvements, particularly in tasks directly related to thermal information. 

In contrast, the RGB-RGB baseline exhibited a relatively high standard deviation. This increased variance resulted from a significant performance drop specifically in thermally dependent sub-tasks compared to other general sub-tasks, creating a performance disparity within the task scenarios.
This suggests that the model can effectively learn behaviors associated with thermal properties even with a limited amount of thermal data. A closer analysis reveals that while the model achieved a high success rate of 83.3\% on tasks involving thermal cues, a slight performance degradation was observed in tasks with low relevance to thermal information but requiring depth perception. In such cases, the model might have relied on the external camera for depth-based operations. In addition, Task1 and Task2 involved everyday objects such as apples, cups, and soda cans, which may be relatively well-represented in large-scale datasets, and this could have contributed to the comparatively higher success rates in these tasks. 

Task 4 is a dynamic task that involves selecting and retrieving an overheated battery from three batteries moving on a conveyor belt. In this task, the proposed model recorded a success rate of 80\%, which is at a similar level to the tasks performed in static environments. This demonstrates that target identification and manipulation based on thermal information can be possible even in dynamic manipulation scenarios where objects are continuously in motion. 

On the other hand, Tasks 3 and 5 focus on perceiving the environment and determining appropriate actions accordingly, rather than directly manipulating thermal-related objects. For instance, in Task 3, the robot selects appropriate actions based on objects such as tea bags or lemons, while in Task 5, it controls the power strip by detecting the overheated state of a hair straightener. These results suggest that even when thermal information is not directly associated with the manipulation target, it can be utilized as a cue in the decision-making stage to drive high task success rates.

In Fig.5, the success rates of thermal-related sub-tasks can be compared. The proposed model showed an average success rate of 82\% in thermal-related sub-tasks, which is an improvement of approximately 40\% compared to the RGB-RGB model (42\%). An analysis of failure cases in Task 3, which showed a relatively low success rate, confirmed that they were caused by the field-of-view  limitation of the wrist camera depending on the picking position. When the back of the scoop was gripped, the wrist camera could not properly capture the color of the beverage, leading to errors in the subsequent pour stage.

 \begin{figure}[t]
   \includegraphics[width=\columnwidth]{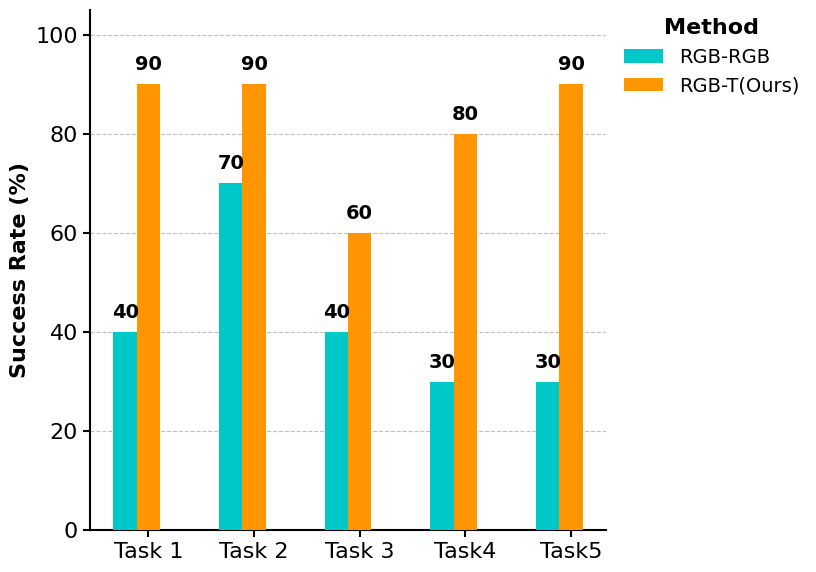}
    \caption{Performance on subtasks requiring thermal perception, including picking up warm water (Task 1), picking up a cold Coke (Task 2), placing an object into the appropriate cup (Task 3), picking up a heated battery (Task 4), and turning off a hair straightener (Task 5).}
    \label{fig:successrate_thermal_perception}
\end{figure}

\subsubsection{Experiment 3: Planning via VLM vs. Flat VLA Models}
This experiment compared the performance of decomposing a task into sub-tasks using a Vision-Language Model (VLM) against learning the entire task as a single policy with a flat VLA. The experiments were conducted under identical environmental conditions, and each task was repeated 10 times. As the Flat VLA had to learn a long-horizon task in an end-to-end manner, its success rate was near zero in most cases, as shown in Table ~\ref{tab:integrated_results}. In contrast, the VLM-based approach, which applied task decomposition, allowed each sub-task to be learned separately, enabling stable execution. For example, it achieved significant success rates in sub-tasks such as the cup and apple manipulation in Task 1 and the hair straightener power control in Task 5. Notably, Task 2 demonstrated the VLM's capability for less direct thermal reasoning. Based on pre-defined sub-task candidates, the VLM performs an environmental analysis using thermal images prior to planning. Consequently, when the analysis revealed that the Coke was warm, the VLM autonomously formulated an adaptive strategy to fetch an ice cup.These results demonstrate that task decomposition through high-level planning is an effective strategy for learning complex manipulation sequences. This is particularly true when using datasets that have not been pre-trained on a large scale, such as thermal imagery, confirming that breaking down a task into sub-tasks can lead to significant performance improvements.

\section{Discussion and Conclusion}
In this study, we introduced ThermoAct, a framework capable of intelligent decision-making based on thermal information. 
By fine-tuning a conventional VLA model with an additional thermal camera input, 
ThermoAct was shown to perform the proposed temperature-based tasks 
in real-world environments.Furthermore, by leveraging a Vision-Language Model (VLM) as a high-level planner, we demonstrated that complex tasks can be decomposed into simpler sub-tasks, which may enable data-efficient learning and stable execution even in the absence of large-scale thermal datasets. This suggests that VLMs may extend beyond mere language understanding to enhance the reasoning capabilities and safety of robotic systems when combined with multimodal sensors. Indeed, in tasks closely associated with thermal information, such as picking up a warm cup or controlling the power of a hair straightener, our model exhibited clear improvements over an RGB-only baseline. This indicates that meaningful performance gains could be achieved with only a small amount of thermal data when combined with a hierarchical planning framework.

However, performance did not improve consistently across all tasks. During the experiments, the robot occasionally performed exploratory actions to locate objects outside the wrist camera’s field of view, and performance degradation was observed in tasks where perceiving the object’s height was critical. This suggests that the robot might have relied on the external camera to estimate the depth required to pick or place objects. In particular, when objects were positioned at lower heights, the robot often hovered above the object, requiring multiple attempts to achieve a correct grasp or failing altogether. These observations imply that a thermal camera alone may not be sufficient for comprehensive environmental perception. Future work should consider (i) a fusion strategy that dynamically adjusts modality contributions depending on task characteristics, (ii)  supplementing spatial awareness with additional inputs, such as a depth sensor. Through such improvements, more stable and generalizable manipulation performance under diverse environmental conditions may be achieved. (iii) scaling up the dataset to encompass a broader range of object categories and temperature conditions to validate scalability and robustness across diverse environments. This study has highlighted both the potential and the limitations of robot manipulation using thermal information, and we anticipate that subsequent research will further expand and refine these findings, ultimately leading to practical real-world applications.

\vspace{15pt}

\begingroup
\tiny  
\setlength\bibitemsep{0pt}  
\printbibliography

@article{cruz2021autonomous,
  title={Autonomous thermal vision robotic system for victims recognition in search and rescue missions},
  author={Cruz Ulloa, Christyan and Prieto S{\'a}nchez, Guillermo and Barrientos, Antonio and Del Cerro, Jaime},
  journal={Sensors},
  volume={21},
  number={21},
  pages={7346},
  year={2021},
  publisher={MDPI}
}

@inproceedings{barros2022cost,
  title={A cost-effective thermal imaging safety sensor for industry 5.0 and collaborative robotics},
  author={Barros, Daniel and Fraga-Lamas, Paula and Fern{\'a}ndez-Caram{\'e}s, Tiago M and Lopes, S{\'e}rgio Ivan},
  booktitle={International Conference on Intelligent Edge Processing in the IoT era},
  pages={3--15},
  year={2022},
  organization={Springer}
}

@article{park2023object,
  title={Object classification system using temperature variation of smart finger device via machine learning},
  author={Park, Heon Ick and Cho, Tae Jin and Choi, In-Geol and Rhee, Min Suk and Cha, Youngsu},
  journal={Sensors and Actuators A: Physical},
  volume={356},
  pages={114338},
  year={2023},
  publisher={Elsevier}
}

@inproceedings{osawa2022material,
  title={Material classification using active temperature controllable robotic gripper},
  author={Osawa, Yukiko and Kase, Kei and Domae, Yukiyasu and Furukawa, Yoshiyuki and Kheddar, Abderrahmane},
  booktitle={2022 IEEE/SICE International Symposium on System Integration (SII)},
  pages={479--484},
  year={2022},
  organization={IEEE}
}

@article{im2025simultaneous,
  title={Simultaneous In-Hand Shape and Temperature Recognition Using Flexible Multilayered Sensor Arrays for Sense-Based Robot Manipulation},
  author={Im, Seong-Min and Park, Byeong-Sun and Jang, Jaehwan and Hong, Sungeun and Nam, Changjoo and Lee, Young Tack and Kim, Min-gu},
  journal={Advanced Sensor Research},
  pages={70004},
  year={2025},
  publisher={Wiley Online Library}
}

@article{kwon2024language,
  title={Language models as zero-shot trajectory generators},
  author={Kwon, Teyun and Di Palo, Norman and Johns, Edward},
  journal={IEEE Robotics and Automation Letters},
  year={2024},
  publisher={IEEE}
}

@article{singh2024malmm,
  title={MALMM: Multi-Agent Large Language Models for Zero-Shot Robotics Manipulation},
  author={Singh, Harsh and Das, Rocktim Jyoti and Han, Mingfei and Nakov, Preslav and Laptev, Ivan},
  journal={arXiv preprint arXiv:2411.17636},
  year={2024}
}

@article{brohan2022rt,
  title={Rt-1: Robotics transformer for real-world control at scale},
  author={Brohan, Anthony and Brown, Noah and Carbajal, Justice and Chebotar, Yevgen and Dabis, Joseph and Finn, Chelsea and Gopalakrishnan, Keerthana and Hausman, Karol and Herzog, Alex and Hsu, Jasmine and others},
  journal={arXiv preprint arXiv:2212.06817},
  year={2022}
}

@article{brohan2023rt,
  title={Rt-2: Vision-language-action models transfer web knowledge to robotic control},
  author={Brohan, Anthony and Brown, Noah and Carbajal, Justice and Chebotar, Yevgen and Chen, Xi and Choromanski, Krzysztof and Ding, Tianli and Driess, Danny and Dubey, Avinava and Finn, Chelsea and others},
  journal={arXiv preprint arXiv:2307.15818},
  year={2023}
}

@article{team2024octo,
  title={Octo: An open-source generalist robot policy},
  author={Team, Octo Model and Ghosh, Dibya and Walke, Homer and Pertsch, Karl and Black, Kevin and Mees, Oier and Dasari, Sudeep and Hejna, Joey and Kreiman, Tobias and Xu, Charles and others},
  journal={arXiv preprint arXiv:2405.12213},
  year={2024}
}

@article{black2024pi_0,
  title={$\pi_0$: A Vision-Language-Action Flow Model for General Robot Control},
  author={Black, Kevin and Brown, Noah and Driess, Danny and Esmail, Adnan and Equi, Michael and Finn, Chelsea and Fusai, Niccolo and Groom, Lachy and Hausman, Karol and Ichter, Brian and others},
  journal={arXiv preprint arXiv:2410.24164},
  year={2024}
}

@article{kim2024openvla,
  title={Openvla: An open-source vision-language-action model},
  author={Kim, Moo Jin and Pertsch, Karl and Karamcheti, Siddharth and Xiao, Ted and Balakrishna, Ashwin and Nair, Suraj and Rafailov, Rafael and Foster, Ethan and Lam, Grace and Sanketi, Pannag and others},
  journal={arXiv preprint arXiv:2406.09246},
  year={2024}
}

@article{pi05,
  author = {{Physical Intelligence} and Black, Kevin and Brown, Noah and Darpinian, James
            and Dhabalia, Karan and Driess, Danny and Esmail, Adnan and Equi, Michael
            and Finn, Chelsea and Fusai, Niccolo and others},
  title  = "{{\ensuremath{\pi_{0.5}}}: A Vision--Language--Action Model with Open-World Generalization}",
  journal= {arXiv preprint arXiv:2504.16054},
  year   = {2025}
}

@article{ma2024survey,
  title={A survey on vision-language-action models for embodied ai},
  author={Ma, Yueen and Song, Zixing and Zhuang, Yuzheng and Hao, Jianye and King, Irwin},
  journal={arXiv preprint arXiv:2405.14093},
  year={2024}
}

@article{sapkota2025vision,
  title={Vision-language-action models: Concepts, progress, applications and challenges},
  author={Sapkota, Ranjan and Cao, Yang and Roumeliotis, Konstantinos I and Karkee, Manoj},
  journal={arXiv preprint arXiv:2505.04769},
  year={2025}
}

@article{hu2023look,
  title={Look before you leap: Unveiling the power of gpt-4v in robotic vision-language planning},
  author={Hu, Yingdong and Lin, Fanqi and Zhang, Tong and Yi, Li and Gao, Yang},
  journal={arXiv preprint arXiv:2311.17842},
  year={2023}
}

@article{shi2025hi,
  title={Hi robot: Open-ended instruction following with hierarchical vision-language-action models},
  author={Shi, Lucy Xiaoyang and Ichter, Brian and Equi, Michael and Ke, Liyiming and Pertsch, Karl and Vuong, Quan and Tanner, James and Walling, Anna and Wang, Haohuan and Fusai, Niccolo and others},
  journal={arXiv preprint arXiv:2502.19417},
  year={2025}
}

@article{yang2025agentic,
  title={Agentic Robot: A Brain-Inspired Framework for Vision-Language-Action Models in Embodied Agents},
  author={Yang, Zhejian and Chen, Yongchao and Zhou, Xueyang and Yan, Jiangyue and Song, Dingjie and Liu, Yinuo and Li, Yuting and Zhang, Yu and Zhou, Pan and Chen, Hechang and others},
  journal={arXiv preprint arXiv:2505.23450},
  year={2025}
}

@inproceedings{guo2024doremi,
  title={Doremi: Grounding language model by detecting and recovering from plan-execution misalignment},
  author={Guo, Yanjiang and Wang, Yen-Jen and Zha, Lihan and Chen, Jianyu},
  booktitle={2024 IEEE/RSJ International Conference on Intelligent Robots and Systems (IROS)},
  pages={12124--12131},
  year={2024},
  organization={IEEE}
}

@inproceedings{kannan2024smart,
  title={Smart-llm: Smart multi-agent robot task planning using large language models},
  author={Kannan, Shyam Sundar and Venkatesh, Vishnunandan LN and Min, Byung-Cheol},
  booktitle={2024 IEEE/RSJ International Conference on Intelligent Robots and Systems (IROS)},
  pages={12140--12147},
  year={2024},
  organization={IEEE}
}

@article{zawalski2024robotic,
  title={Robotic control via embodied chain-of-thought reasoning},
  author={Zawalski, Micha{\l} and Chen, William and Pertsch, Karl and Mees, Oier and Finn, Chelsea and Levine, Sergey},
  journal={arXiv preprint arXiv:2407.08693},
  year={2024}
}

@article{zhen20243d,
  title={3d-vla: A 3d vision-language-action generative world model},
  author={Zhen, Haoyu and Qiu, Xiaowen and Chen, Peihao and Yang, Jincheng and Yan, Xin and Du, Yilun and Hong, Yining and Gan, Chuang},
  journal={arXiv preprint arXiv:2403.09631},
  year={2024}
}

@article{yu2025forcevla,
  title={ForceVLA: Enhancing VLA Models with a Force-aware MoE for Contact-rich Manipulation},
  author={Yu, Jiawen and Liu, Hairuo and Yu, Qiaojun and Ren, Jieji and Hao, Ce and Ding, Haitong and Huang, Guangyu and Huang, Guofan and Song, Yan and Cai, Panpan and others},
  journal={arXiv preprint arXiv:2505.22159},
  year={2025}
}

@article{ashqar2024leveraging,
  title={Leveraging Multimodal Large Language Models (MLLMs) for Enhanced Object Detection and Scene Understanding in Thermal Images for Autonomous Driving Systems},
  author={Ashqar, Huthaifa I and Alhadidi, Taqwa I and Elhenawy, Mohammed and Khanfar, Nour O},
  journal={Automation},
  volume={5},
  number={4},
  pages={508--526},
  year={2024},
  publisher={MDPI}
}

@article{zhang2025vtla,
  title={Vtla: Vision-tactile-language-action model with preference learning for insertion manipulation},
  author={Zhang, Chaofan and Hao, Peng and Cao, Xiaoge and Hao, Xiaoshuai and Cui, Shaowei and Wang, Shuo},
  journal={arXiv preprint arXiv:2505.09577},
  year={2025}
}

@article{zhao2025vlas,
  title={Vlas: Vision-language-action model with speech instructions for customized robot manipulation},
  author={Zhao, Wei and Ding, Pengxiang and Zhang, Min and Gong, Zhefei and Bai, Shuanghao and Zhao, Han and Wang, Donglin},
  journal={arXiv preprint arXiv:2502.13508},
  year={2025}
}

@article{huang2025tactile,
  title={Tactile-VLA: Unlocking Vision-Language-Action Model's Physical Knowledge for Tactile Generalization},
  author={Huang, Jialei and Wang, Shuo and Lin, Fanqi and Hu, Yihang and Wen, Chuan and Gao, Yang},
  journal={arXiv preprint arXiv:2507.09160},
  year={2025}
}

@article{jones2025beyond,
  title={Beyond sight: Finetuning generalist robot policies with heterogeneous sensors via language grounding},
  author={Jones, Joshua and Mees, Oier and Sferrazza, Carmelo and Stachowicz, Kyle and Abbeel, Pieter and Levine, Sergey},
  journal={arXiv preprint arXiv:2501.04693},
  year={2025}
}

@article{ko2023vision,
  title={Vision-based interaction force estimation for robot grip motion without tactile/force sensor},
  author={Ko, Dae-Kwan and Lee, Kang-Won and Lee, Dong Han and Lim, Soo-Chul},
  journal={Expert Systems with Applications},
  volume={211},
  pages={118441},
  year={2023},
  publisher={Elsevier}
}

@article{lee2024dextouch,
  title={Dextouch: Learning to seek and manipulate objects with tactile dexterity},
  author={Lee, Kang-Won and Qin, Yuzhe and Wang, Xiaolong and Lim, Soo-Chul},
  journal={IEEE Robotics and Automation Letters},
  year={2024},
  publisher={IEEE}
}

@article{hao2025tla,
  title={Tla: Tactile-language-action model for contact-rich manipulation},
  author={Hao, Peng and Zhang, Chaofan and Li, Dingzhe and Cao, Xiaoge and Hao, Xiaoshuai and Cui, Shaowei and Wang, Shuo},
  journal={arXiv preprint arXiv:2503.08548},
  year={2025}
}

@misc{GoogleBlogGemini2,
  author = {{Google Developers}},
  title = {{Gemini 2.0: Flash, Flash-Lite and Pro}},
  howpublished = {\url{https://developers.googleblog.com/en/gemini-2-family-expands/}},
  month = feb,
  year = {2025},
  note = {Accessed: 20250911},
}

@article{hu2022lora,
  title={Lora: Low-rank adaptation of large language models.},
  author={Hu, Edward J and Shen, Yelong and Wallis, Phillip and Allen-Zhu, Zeyuan and Li, Yuanzhi and Wang, Shean and Wang, Lu and Chen, Weizhu and others},
  journal={ICLR},
  volume={1},
  number={2},
  pages={3},
  year={2022}
}
\endgroup
\addtolength{\textheight}{-12cm}   

\end{document}